\def\year{2022}\relax
\documentclass[letterpaper]{article} 
\usepackage{aaai22}  
\usepackage{times}  
\usepackage{helvet}  
\usepackage{courier}  
\usepackage[hyphens]{url}  
\usepackage{graphicx} 
\usepackage{natbib}  
\usepackage{caption} 
\DeclareCaptionStyle{ruled}{labelfont=normalfont,labelsep=colon,strut=off} 
\usepackage{algorithm}
\usepackage{algorithmic}

\usepackage{newfloat}
\usepackage{listings}
\floatstyle{ruled}
\newfloat{listing}{tb}{lst}{}
\floatname{listing}{Listing}
\title{Exploiting the Potential of Datasets: A Data-Centric Approach for Model Robustness}
\author{
    Yiqi Zhong,
    Lei Wu,
    Xianming Liu\thanks{Corresponding author},
    Junjun Jiang
}
\affiliations{

    Advanced Imaging and Intelligent Analysis Lab\\
    School of Computer Science and Technology\\
    Harbin Institute of Technology, Harbin, China\\
    \{21s003117, 21s003031\}@stu.hit.edu.cn, \{csxm, jiangjunjun\}@hit.edu.cn
}

\usepackage{bibentry}

\usepackage{hyperref}
\hypersetup{
    colorlinks=true,
    linkcolor=black,
    urlcolor=magenta,
    citecolor=black,
}
\usepackage{bm}
\usepackage{amssymb}
\usepackage{amsmath}
\usepackage{booktabs}
\usepackage{multirow}
\usepackage{xcolor}
\usepackage[normalem]{ulem} 
\newcommand\hl{\bgroup\markoverwith
  {\textcolor{yellow}{\rule[-.5ex]{2pt}{2.5ex}}}\ULon}

\begin{document}

\maketitle

\begin{abstract}
Robustness of deep neural networks (DNNs) to malicious perturbations is a hot topic in trustworthy AI. Existing techniques obtain robust models given fixed datasets, either by modifying model structures, or by optimizing the process of inference or training. While significant improvements have been made, the possibility of constructing a high-quality dataset for model robustness remain unexplored. Follow the campaign of data-centric AI launched by Andrew Ng, we propose a novel algorithm for dataset enhancement that works well for many existing DNN models to improve robustness. Transferable adversarial examples and 14 kinds of common corruptions are included in our optimized dataset. In the data-centric robust learning competition hosted by Alibaba Group and Tsinghua University, our algorithm came third out of more than 3000 competitors in the first stage while we ranked fourth in the second stage. Our code is available at \href{https://github.com/hncszyq/tianchi_challenge}{https://github.com/hncszyq/tianchi\_challenge}.
\end{abstract}
\section{1. Introduction}

Deep learning has set off a revolution in artificial intelligence research and has made remarkable achievements in many fields such as medical diagnosis, autonomous driving, large-scale decision making, etc. However, it's been proved that DNNs are vulnerable to adversarial examples \cite{szegedy2013intriguing}, which are clean samples with imperceptible perturbations that cause a model to make mistakes, posing a serious threat to AI security. For adversarial defense, existing work either modify the model structures themselves, or optimize the process of inference or training, among which adversarial training \cite{madry2017towards, zhang2019theoretically} proves to be the most effective strategy.

Some works point out that DNNs are also susceptible to common corruptions that widely exist in real-world application scenarios \cite{LinfengZhang2020AuxiliaryTT, DanHendrycks2019BenchmarkingNN}. These corruptions stem from geometric variations of cameras caused by rotation and translation, or some  environmental factors like rain, snow, noises, etc. The techniques towards robustness to common corruptions are mainly focus on data augmentation \cite{DanHendrycks2020AugMixAS} and auxiliary training \cite{zheng2016improving, LinfengZhang2020AuxiliaryTT}. 

Numerous works have tried to improve robustness of DNN models. However, none of them considered engineering the original dataset to make it more suitable for training robust models, leaving the potential of datasets unexplored. The encouraging result of data-centric AI competition launched by Andrew Ng (\href{https://https-deeplearning-ai.github.io/data-centric-comp/}{url}) tells us that it's time to move from model-centric approach to data-centric approach and design reliable, effective, systematic data to furthur stimulate the potential of deep learning. 

In this paper, we propose a data-centric algorithm for dataset enhancement to train robust models. Based on the following two conclusions: 1) a DNN model achieves optimal performance when its testing set follow the same data distribution as its training set; 2) maliciously perturbed data and benign data come from different distributions \cite{song2017pixeldefend, samangouei2018defense}, we believe that the optimized training set should also contain adversarial examples and corrupted samples drawn from the corresponding distribution. Therefore, the basic framework of our algorithm is quit simple --- we add adversarial perturbations and common corruptions to some randomly selected samples in the raw training set. We make use of transferable adversarial examples and as many as 14 kinds of corruptions to further improve the effectiveness. Note that this is different from the techniques exploring extra data which usually result in larger datasets, we keep the number of training samples unchanged. Our algorithm is proposed for participating in the data-centric robust learning competition hosted by Alibaba Group and Tsinghua University, in which we beat over 3,000 participants and won the 3rd and 4th place in stage one and stage two, respectively. Our main contributions are as follows:

\begin{itemize}
    \item We propose a simple but effective algorithm to improve deep model’s robustness from the perspective of data. This brand new and promising data-centric view largely enrich the research community.
    \item We study the robustness of DNN models in a challenging but practical setting --- adversarial examples and common corruptions both exist in the testing phase while most of the previous work consider them separately.
    \item The competition and experimental results demonstrated the effectiveness of our algorithm, indicating that the data-centric strategy is feasible for model robustness. 
\end{itemize}

\section{2. Related Works}
\subsection{Model Robustness}
Robustness of DNNs to the perturbations on model inputs is of great concern in trustworthy AI. There are two kinds of perturbations studied in the literature, one is adversarial perturbation \cite{szegedy2013intriguing} which is a small perturbation that can drastically change the network output while being quasi-imperceptible to humans, and another is common corruption such as rain, snow, Gaussian noise, etc. 

The research community has made great efforts to improve model robustness. To defense against adversarial examples, one of the most effective strategies is adversarial training --- train a model in an adversarial fashion that continuously generating adversarial examples and then minimizing the loss on these samples \cite{madry2017towards, zhang2019theoretically}. In addition, There are also some works that pursue robust model structures by leveraging ensemble strategies \cite{lu2021armoured}, NAS \cite{hosseini2021dsrna}, or some well-designed modules for denoising \cite{xie2019feature}, purifying \cite{shi2020online}, and malicious sample rejection \cite{cohen2020detecting}. For common corruptions, existing works mainly focus on optimizing the learning strategies \cite{hendrycks2019using, zheng2016improving} or data augmentation \cite{devries2017improved}. Note that data augmentation can be seen as a data-centric algorithm, but it often result in a larger training set while our algorithm does not.

The above strategies, while effective, are based on fixed datasets or extra training samples. From a complementary perspective, in this work, we show that it's possible to effectively improve model robustness simply by improving existing datasets while without increasing the amount of data.

\subsection{Data-Centric AI}
Data-centric AI stems from a competition launched by Andrew Ng (\href{https://https-deeplearning-ai.github.io/data-centric-comp/}{url}). Different from previous competitions that pursue high-performance models with fixed dataset, this competition fix the models and pursue a high-quality dataset by fixing incorrect labels, applying data augmentations, etc. In this work, we follow a stricter requirement --- improve a dataset without increasing the number of samples. For model robustness, previous work focus almost exclusively on models, and now it's the time to exploit the potential of datasets.

\section{3. Proposed Method}
\label{sec:section3}

In this section, we present our data-centric algorithm for model robustness. We first formalize our optimization goal.

\subsection{Problem Formulation}

Given a training set $\mathcal{D}=\left\{(x_1, y_1),\cdots,(x_n, y_n)\right\}$ consisting of $n$ image-label pairs and a DNN-based classifier $f(\theta;x):\mathbb{R}^d\rightarrow\mathbb{R}^k$ with parameter $\theta$, a standard scheme to train model $f$ is empirical risk minimization (ERM). Let $J(\theta;x,y)$ be the loss function of $\theta$ with input $x$ and one-hot label $y$. Usually, $J$ can be KL divergence, i.e.,
$$J(\theta;x,y)=\text{KL}(f(\theta;x)\|y).$$
The objective of ERM on training set $\mathcal{D}$ can be formulated as follows:
$$ERM(\mathcal{D})=\mathop{\arg\min}\limits_{\theta\in\Theta}\mathbb{E}_{(x,y)\sim\mathcal{D}}[J(\theta;x,y)],$$
where $\Theta$ is the parameter space. Suppose there is a robustness metric $R(f, \theta)$ and a larger value indicates a more robust classifier $f$ with parameter $\theta$, our goal is to develop a data-centric algorithm $\mathcal{A}$  such that $R\left(f, ERM(\mathcal{A}(\mathcal{D}))\right)$ can be maximized and $|\mathcal{A}(\mathcal{D})|$ is not greater than $|\mathcal{D}|$, where $\mathcal{A}(\mathcal{D})$ is the enhanced dataset used to train robust models. We will give the definition of $R(f, \theta)$ in Sec. 4.

\subsection{Our Data-centric Algorithm}

\begin{figure}[t]
    \centering
    \includegraphics[width=1.0\linewidth,trim=60 0 65 0]{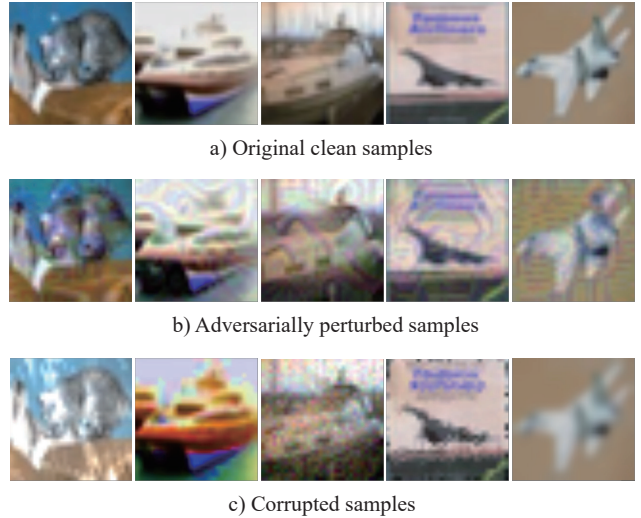}
    \caption{Comparison of our generated perturbed samples and the original clean samples.}
    \label{fig:example}
\end{figure}

For a DNN model, to achieve optimal performance in the testing phase, the testing set should follow the same data distribution as our training set. However, numerous works have shown that maliciously perturbed data and benign data come from different distributions, which result in poor performance of DNNs under attack \cite{song2017pixeldefend, samangouei2018defense}. This inspires us that our training set should also contain perturbed samples drawn from the corresponding distribution. 

In summary, our data-centric algorithm $\mathcal{A}$ can be described as follows. We randomly split the original training set $\mathcal{D}$ into three parts $\mathcal{D}_1$, $\mathcal{D}_2$, and $\mathcal{D}_3$ in a ratio of $\alpha_1:\alpha_2:\alpha_3$. For $\mathcal{D}_1$, we keep the original samples unchanged; for $\mathcal{D}_2$, we adversarially perturb each sample; for $\mathcal{D}_3$, we apply a random kind of common corruption to each sample. Finally,  $\cup_{i=1}^3\mathcal{D}_i$ is our optimized training set. 

Next, we introduce our technical details about adversarial perturbations and common corruptions.

\textbf{Adversarial perturbation}. A successful data-centric algorithm should be a plug and play approach that works well for any model $f$. Therefore, the generated adversarial perturbations should be transferable, which means they can fool any unknown models that perform the same task. In general, the process of generating adversarial perturbations can be formulated as follows\footnote{Note that this is the formulation of $l_\infty\text{-bounded}$ PGD attack. For $l_2\text{-bounded}$ PGD attack, we should replace $\text{sign}(\cdot)$ with $\text{norm}(\cdot)$, where $\text{norm}(v)=\frac{v}{\|v\|_2}$.}:
\begin{equation}
\delta_{i+1}=\text{Proj}_
\epsilon\left(\delta_i+\alpha\cdot\text{sign}(\nabla_{\delta_i} J(\theta; x+\delta_i, y))\right), \notag
\end{equation}
where $\delta_0$ is a randomly initialized perturbation and $\text{Proj}_\epsilon(\cdot)$ projects the current $\delta_i$ into the $l_p$ norm ball with radius $\epsilon$. There are many effective techniques in the literature to enhance the transferability of adversarial examples, e.g.,
\begin{itemize}
   \item \textit{Momentum update} \cite{dong2018boosting}, which integrates a momentum term into the calculation of gradients to stabilize update directions and avoid poor local optima:
   \begin{equation}
   \begin{aligned}
   &g_i=\mu\cdot g_{i-1}+\frac{\nabla_{\delta_i} J(\theta;x+\delta_i,y)}{\|\nabla_{\delta_i} J(\theta;x+\delta_i,y)\|_1},\\
   &\delta_{i+1}=\text{Proj}_\epsilon(\delta_i+\alpha\cdot\text{sign}(g_i)).
   \end{aligned}
   \end{equation}
   \item \textit{Gradient smoothing} \cite{dong2019evading}, which applies Gaussian smoothing to the gradients to weaken their correlation with a particular model:
   \begin{equation}
   \begin{aligned}
   \delta_{i+1}=\text{Proj}_\epsilon(&\delta_i+\alpha\cdot\text{sign}(\\ &W\ast\nabla_{\delta_i}J(\theta; x+\delta_i, y))),
   \end{aligned}
   \end{equation}
   where $W$ is the Gaussian kernel.
   \item \textit{Input diversification} \cite{xie2019improving}, which applies random transformations to the input images at each iteration that can be seen as a special kind of data augmentation to avoid overfitting:
   \begin{equation}
   \begin{aligned}
   \delta_{i+1}=\text{Proj}_\epsilon(&\delta_i+\alpha\cdot\text{sign}(\\ &\nabla_{\delta_i}J(\theta; T(x+\delta_i,p), y))),
   \end{aligned}
   \end{equation}
   where $T(x+\delta_i,p)$ is the randomly transformed sample.
   \item \textit{Logit loss} \cite{zhao2021success}, which can avoid the vanishing gradient problem caused by cross entropy loss:
   \begin{equation}
   \begin{aligned}
   J(\theta;x,y)=-z^t,\\
   t=\arg\max_iy^i.
   \end{aligned}
   \end{equation}
   where $z$ is the output of the logit layer.
   \item \textit{Model ensemble} \cite{dong2018boosting}, which fuses the logits of multiple models together to get the final output:
   \begin{equation}
   Z_{ensemble}=\frac{1}{|F|}\sum_{f\in F}z_f
   \end{equation}
   where $F$ is the model set, $z_f$ is the logit of model $f$ and $|F|$ is the cardinality of $F$.
\end{itemize}

Following  \cite{zhao2021success}, we combine $\text{(1)}\sim\text{(5)}$ together to generate highly transferable adversarial perturbations. The iterative formulas are as follows:
\begin{equation}
\begin{aligned}
  &\delta_i = T(x+\delta_i, p)-x,\\
  &J(\theta_F;x+\delta_i,y)=-Z_{ensemble}^t,\\
  &g_i=\mu\cdot g_{i-1}+\frac{\nabla_{\delta_i}J(\theta_F;x+\delta_i,y)}{\|\nabla_{\delta_i}J(\theta_F;x+\delta_i,y\|_1},\\
  &\delta_{i+1}=\text{Proj}_\epsilon(\delta_i+\alpha\cdot\text{sign}(W\ast g_i)).
\end{aligned}
\end{equation}

To further obtain more diverse perturbation patterns, we generate both $l_2$-bounded and $l_\infty$-bounded perturbations for each sample. Next, we add them together to get the final adversarial perturbation. In Fig. \ref{fig:example}b, we show some instances of our generated adversarial examples.

\textbf{Common corruption}. Using the imgaug library \cite{imgaug}, we implement 14 kinds of common corruptions including rain, snow, frost, Gaussian noise, elastic transformation, etc. For each sample in $\mathcal{D}_3$, we randomly select one of the 14 methods to corrupt it. In Fig. \ref{fig:example}c, we show some instances of our corrupted samples.

\section{4. Experiments}

\begin{table*}[t]
	\caption{Performance comparison of models trained on original CIFAR-10 and our optimized CIFAR-10. $ACC(\cdot)$ stands for the classification rate on some subdataset and $R(f,\theta)$ stands for the robustness score of model $f$ with parameter $\theta$.}
	\centering
	\begin{tabular}{lcccccc}
		\toprule
		\multirow{2}{*}{}
		&\multicolumn{6}{c}{\textbf{Performances of models trained on original CIFAR-10 (\%)}} \\
		
		\cline{2-7} 
		\specialrule{0em}{1pt}{1pt}
		&ResNet50 &WideResNet &PreactResNet18 &DenseNet121 &VGG16 &MobileNetV2 \\
		\midrule
		$ACC(\mathcal{P}_{ori})$ 
		&98.64 &99.28 &98.66 &98.63 &98.66 &90.91 \\
		$ACC(\mathcal{P}_{adv})$          
		&32.73 &34.28 &33.75 &31.84 &37.17 &43.25 \\
		$ACC(\mathcal{P}_{cor})$
		&55.12 &61.32 &62.78 &56.71 &64.17 &56.67\\
		\cline{1-7} 
		\specialrule{0em}{1pt}{1pt}
		$R(f,\theta)$
		&62.16 &64.96 &65.06 &62.39 &66.67 &63.61\\
		\toprule
		\multirow{2}{*}{}
		&\multicolumn{6}{c}{\textbf{Performances of models trained on optimized CIFAR-10 (\%)}} \\
		
		\cline{2-7} 
		\specialrule{0em}{1pt}{1pt}
		&ResNet50 &WideResNet &PreactResNet18 &DenseNet121 &VGG16 &MobileNetV2 \\
		\midrule
		$ACC(\mathcal{P}_{ori})$ 
		&92.75 &96.30 &93.30 &95.04 &94.94 &84.74 \\
		$ACC(\mathcal{P}_{adv})$          
		&60.52 &62.42 &58.34 &61.50 &58.86 &46.99 \\
		$ACC(\mathcal{P}_{cor})$
		&83.36 &89.58 &83.93 &85.54 &86.93 &72.58\\
		\cline{1-7} 
		\specialrule{0em}{1pt}{1pt}
		$R(f,\theta)$
		&\textbf{78.88} &\textbf{82.77} &\textbf{78.52} &\textbf{80.69} &\textbf{80.24} &\textbf{68.10}\\
		\bottomrule
	\end{tabular}
	\label{tab:comparison}
\end{table*}

In this section, we will empirically demonstrate the effectiveness of our designed data-centric algorithm $\mathcal{A}$, which is proposed for participating in the data-centric robust learning competition hosted by Alibaba Group and Tsinghua University as one of the series of AI Security Challengers Program (\href{https://tianchi.aliyun.com/competition/entrance/531939/introduction}{url}). We first introduce this competition as well as the robustness metric to evaluate an algorithm. 

\subsection{Data-Centric Robust Learning Competition}
 In this competition, we need to optimize the CIFAR-10 dataset using our data-centric algorithm $\mathcal{A}$. Based on this dataset, we are able to train some robust models. To evaluate the robustness of these models, the competition constructed a private testing set $\mathcal{P}=\{\mathcal{P}_{ori},\mathcal{P}_{adv},\mathcal{P}_{cor}\}$ based on CIFAR-10. $\mathcal{P}$ consists of three subdatasets which contain clean samples, adversarial examples and corrupted samples respectively\footnote{The clean samples in $\mathcal{P}$ are not necessary selected from the original CIFAR-10 dataset.}. For a particular model $f$ with parameter $\theta$, its robustness $R(f, \theta)$ is defined as the classification rate on $\mathcal{P}$, which can be formulated as follows\footnote{Here, the label $y$ and the output of $f$ are all scalars representing the class indexs, which are different from the definitions in Seq. 3.}:
 \begin{equation}
     R(f, \theta)=\frac{1}{|\mathcal{P}|}\left(\sum_{\mathcal{P}_i\in\mathcal{P}}\frac{1}{|\mathcal{P}_i|}\sum_{(x_i,y_i)\in\mathcal{P}_i}\bm{1}(f(\theta,x_i)=y_i)\right).
 \end{equation}

This competition consists of two stages. In each stage, we are given several DNN models and we need to train these models on our optimized dataset. The trained models are subsequently submitted to the competition platform. Finally, our score is calculated as the mean of $R$ for each model based on Eq. (7). In the first stage, the models to be trained are ResNet50 \cite{he2016deep} and DenseNet121 \cite{huang2017densely}. The score of the baseline dataset is 75.23. We came third out of 3691 participants with a score of 98.96. In the second stage, the models to be trained are WideResNet \cite{zagoruyko2016wide} and PreactResNet18 \cite{he2016identity}. The models are evaluated on a different private testing set and the baseline is 63.31. We came fourth out of 50 participants with a score of 85.19.

\subsection{Comparison With the Baseline}

Exploiting data-centric AI for model robustness is a brand new idea with no competitive method in the literature. Therefore, our baseline algorithm for comparison is simply the identity map $\mathcal{I}$, i.e., the original CIFAR-10 is the baseline dataset to train non-robust models. Note that the number of training samples contained in our optimized dataset should not exceed 50,000 during the competition, which is less than the total number of samples in CIFAR-10 (includes training set and testing set). As what we did in the competition, we randomly remove 10,000 samples from CIFAR-10 before our experiments. 

Using the proposed algorithm $\mathcal{A}$ described in Sec. 3, we generate our optimized training set based on the original CIFAR-10. In practice, the split ratio $\alpha$ is set as $0:1:4$, which yields the best performance in the competition. This ratio indicates that there are no clean samples in our optimized training set. This is because the proportion of clean samples in the private testing set $\mathcal{P}$ is relatively small (4.55\% in the second stage), and also, the perturbed samples contain some information about the distribution of clean samples. We generate transferable adversarial examples through 300 iterations. The hyper-parameters setting in the experiments is: $\epsilon=8$, $\alpha=2/255$ for $l_\infty$-bounded perturbations and $\epsilon=1.0$, $\alpha=0.025$ for $l_2$-bounded perturbations. Other hyper-parameters are consistent with those in \cite{zhao2021success}. To implement model ensemble, we select 5 common DNN models, including ResNet50, WideResNet, DenseNet121, VGG16 \cite{simonyan2014very} and MobileNetV2 \cite{sandler2018mobilenetv2}. Among them, ResNet50 and DenseNet121 have been demonstrated to be the best choices to generate transferable adversarial examples.

We trained 6 different models on both the original CIFAR-10 and our optimized CIFAR-10. The private testing sets of the competition have been released. Therefore, we simply use the testing set of the second stage to evaluate the performances of our trained models. The classification rates $ACC$ on three subdatasets and the robustness score $R(f,\theta)$ of each model are shown in Tab.  \ref{tab:comparison}. The models trained on our optimized dataset show significant improvements in terms of robustness, and the classification rates on both adversarial and corrupt samples increase by more than 20\%, which proves the effectiveness and general applicability of our algorithm. In practice, we can trade off the performance loss on clean samples and the robustness to perturbations by adjusting the split ratio $\alpha$.



\subsection{Limitation and Future Work}

Although our algorithm effectively improves the robustness of models to adversarial and corrupted samples, the magnitude of improvements are not comparable to current state-of-the-art models-centric techniques. On one hand, our study is based on a more challenging setting --- both adversarial examples and corrupted samples exist in the testing set. On the other hand, when the number of samples in the training set cannot be increased, it's difficult to carry enough information about the distributions of clean samples and perturbed samples. While only limited performance gains can be obtained by optimizing the datasets alone, it will be interesting to explore whether further breakthroughs can be achieved by combining data-centric approaches with state-of-the-art model-centric approaches, which is a promising future work.

\section{5. Conclusion}

In this paper, we show the possibility of constructing a high-quality dataset for model robustness by presenting a novel data-centric algorithm. Our competition and experimental results demonstrate the effectiveness and general applicability of the algorithm. Data-centric AI is a promising approach in the field of model robustness, and we believe that more encouraging results can be achieved by combining it with existing state-of-the-art model-centric techniques.

\section{6. Acknowledgments}
This work was supported by  National Natural Science Foundation of China under Grants 61922027, 6207115 and 61932022. We thank the security AI challenger program launched by Alibaba Group and Tsinghua University. We'd also like to thank Xiong Zhou, Chenyang Wang and Xingyu Hu for their kindness help during the competition.

\bibliography{robust_learning}

\end{document}